\documentclass[letterpaper]{article} 
\usepackage{aaai25}  
\usepackage{times}  
\usepackage{helvet}  
\usepackage{courier}  
\usepackage[hyphens]{url}  
\usepackage{graphicx} 
\usepackage{natbib}  
\usepackage{caption} 
\usepackage{algorithm}
\usepackage{algorithmic}
\usepackage{booktabs}
\usepackage[table]{xcolor}
\usepackage{graphicx}
\usepackage{amsmath}
\usepackage{amssymb}
\usepackage{amsthm}
\theoremstyle{Theorem}
\newtheorem{thm}{Theorem}
\newtheorem{ass}{Assumption}
\newtheorem{lem}{Lemma}
\newtheorem{corollary}{Corollary}[thm]
\usepackage{caption}
\usepackage{pgf}
\usepackage{fp}
\newcommand{\numstd}[2]{#1$\pm$\FPeval{\result}{round(#2,1)}\result}

\usepackage{newfloat}
\usepackage{listings}
\DeclareCaptionStyle{ruled}{labelfont=normalfont,labelsep=colon,strut=off} 
\floatstyle{ruled}
\newfloat{listing}{tb}{lst}{}
\floatname{listing}{Listing}
\title{LCGC: Learning from Consistency Gradient Conflicting for Class-Imbalanced Semi-Supervised Debiasing}
\author{
    Weiwei Xing\textsuperscript{\rm 1}, Yue Cheng \textsuperscript{\rm 1}, Hongzhu Yi \textsuperscript{\rm 1},\\
    Xiaohui Gao\textsuperscript{\rm 2},
    Xiang Wei \textsuperscript{\rm 1} \thanks{Corresponding author.},
    Xiaoyu Guo \textsuperscript{\rm 1},
    Yuming Zhang \textsuperscript{\rm 3},
    Xinyu Pang \textsuperscript{\rm 4} 
}
\affiliations{
    \textsuperscript{\rm 1}School of Software Engineering, Beijing Jiaotong University\\
    \textsuperscript{\rm 2}School of Automation, Northwestern Polytechnical University\\
    \textsuperscript{\rm 3}School of Computing, Newcastle University\\
    \textsuperscript{\rm 4}School of Science, Chongqing University of Posts and Telecommunications\\
    \{wwxing, yuecheng, hongzhuyi, xiangwei, guoxiaoyu\}@bjtu.edu.cn, gaitxh@foxmail.com\\
    y.zhang361@newcastle.ac.uk, candypxy@163.com
}

\usepackage{bibentry}

\begin{document}

\maketitle

\begin{abstract}
Classifiers often learn to be \textbf{\textit{biased}} corresponding to the class-imbalanced dataset, especially under the semi-supervised learning (SSL) set. While previous work tries to appropriately re-balance the classifiers by subtracting a class-irrelevant image's logit, but lacks a firm theoretical basis. We theoretically analyze why exploiting a baseline image can refine pseudo-labels and prove that the black image is the best choice. We also indicated that as the training process deepens, the pseudo-labels before and after refinement become closer. Based on this observation, we propose a debiasing scheme dubbed \textbf{LCGC}, which Learning from Consistency Gradient Conflicting, by encouraging biased class predictions during training. We intentionally update the pseudo-labels whose gradient conflicts with the debiased logits, representing the optimization direction offered by the over-imbalanced classifier predictions. Then, we debiased the predictions by subtracting the baseline image logits during testing. Extensive experiments demonstrate that \textbf{LCGC} can significantly improve the prediction accuracy of existing CISSL models on public benchmarks.
\end{abstract}


\section{Introduction}
\label{sec:Intro}
The predictions of an ideal unbiased classifier are attributed to class-relevant features of the input images. In real cases, however, such an ideal classifier can hardly be obtained due to the dataset selection biases and imbalances, the complexity of the real world, or other peculiarities \cite{Bengio2020A}. That's why classifiers trained on a class-imbalanced set may fail to make trustworthy predictions, which is even more pronounced in semi-supervised learning (SSL) \cite{chen2023softmatch}. 

\begin{figure}[!htbp]
    \centering
    \includegraphics[width=0.91\columnwidth]{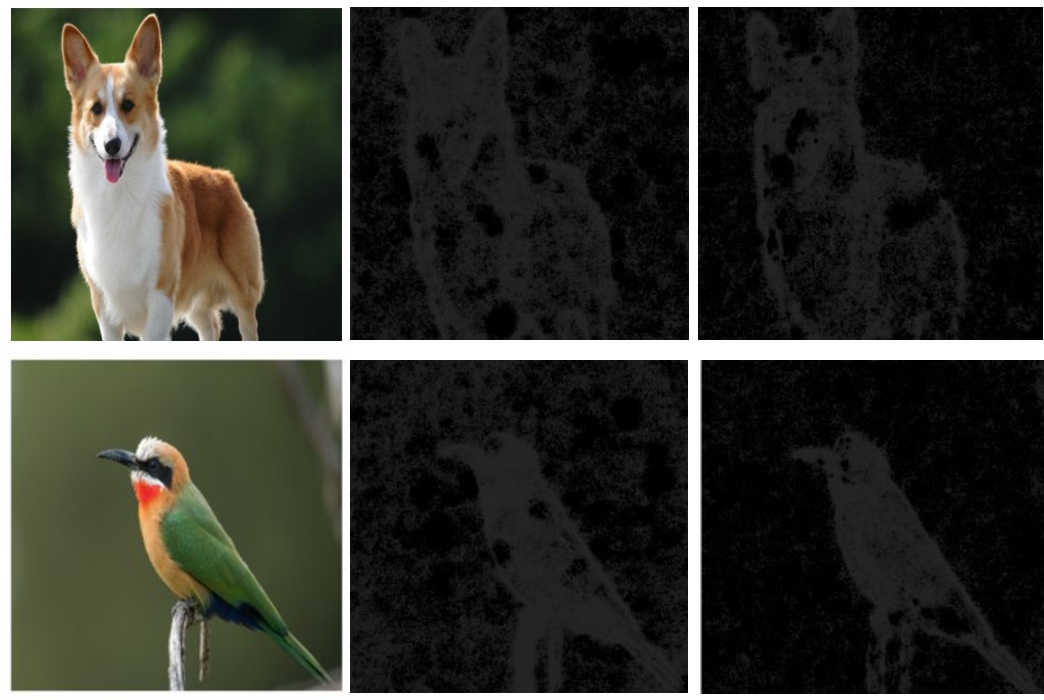}
    \caption{Visualization of sensitive maps produced by the FixMatch model and CDMAD at the image. Left-to-right: original input image, visualization of sensitive maps produced by the FixMatch, CDMAD. The lightness of the sensitive map indicates how much attention the models pay to a particular area of the input image. The sensitive maps obtained by CDMAD have less noise.}
    \label{fig:visualization}
\end{figure}

Essentially, when there is training with imbalanced data, the accuracy of the model drops mainly caused by the spurious correlation between the irrelevant features and category labels, i.e. the classifier output logits are biased attributes towards the majority class of class-dependent features. Taking the task of recognizing animal categories as an example, as shown in Figure \ref{fig:visualization}, the sensitive map obtained after vanilla FixMatch gives certain attention to class-irrelevant background features, while the debias model directs its attention more towards class-relevant features. This results in the latter having less background noise.

To address this issue, a few algorithms \cite{Oh_2022_CVPR, Wang_2022_CVPR, schmutz2023dont, Lee_2024_CVPR} have been proposed for classifier debiasing under class-imbalanced SSL (CISSL) settings. The basic idea of classifier debiasing is to regularize the response bias generated by the biased classification head to achieve the balance effect \cite{Wei_2023_CVPR, wang2023imbalanced, zhu2024estimating}. Such a training strategy alleviates the bias caused by the imbalanced dataset to a certain extent. However, they need to design complex network variants, such as adding classification heads \cite{NEURIPS2021_3953630d}, designing causal balancers \cite{Wang_2022_CVPR, Li_Tao_Han_Zhan_Ye_2024, 10446881}, etc. An immediate idea is whether there is a way to measure the bias of the classifier in a minimally expensive way and to perform post-hoc logit adjustment on the response bias generated by the biased classification head. Class-Distribution-Mismatch-Aware Debiasing (CDMAD) does this job well with a solid color image \cite{Lee_2024_CVPR}. However, what we care about \textit{is there any theoretical evidence to support that a solid color image can debias the classifier logits during training?}

In this paper, our theoretical analysis shows that the CDMAD enhances the balance of the base SSL model implicitly utilizing the integrated gradient flow. It is then natural to ask: \textit{are there any further debiasing methods that can improve the performance of the classifier?} If so, we still expect the new proxy method to be amenable to providing insights for classifier debias and a wider spectrum of architectures. To this end, we propose a novel debiasing model \textbf{LCGC}, by \textbf{\underline{L}}earning from \textbf{\underline{C}}onsistency \textbf{\underline{G}}radient \textbf{\underline{C}}onflicting, to overcome the improper training bias from class-imbalanced datasets. In particular, we first update the pseudo-labels whose gradient conflicts with the baseline image refined logits, which is represented as the optimization direction offered by the over-imbalanced classifier predictions. Here, we employ Kullback-Leibler (KL) divergence to measure the consistency between the refinement and original pseudo-labels logits. Encouraging the conflict between them, we train an \textbf{\textit{over-biased}} classifier neural network. Then, we expected to debias the predictions by subtraction the baseline image logits during testing. 

Our work makes three major contributions. 1) We theoretically analyze that a baseline image enhances the base SSL model by implicitly utilizing the integrated gradient flow. 2) We propose a simple debiased scheme by learning consistency gradient conflicting. 3) We validated the effectiveness of LCGC on four benchmark datasets in both scenarios where the class distributions of the labeled and unlabeled sets either match or mismatch.

\section{Related Work}
\label{sec:Related}
\subsection{Classifier Debias}
\label{sec:cd}
For the classifier debiasing problem, the mainstream methods involve adjustments through re-weighting \cite{Zhang_2021_CVPR} and re-sampling \cite{NEURIPS2021_3953630d}, as well as the lately prevalent logits adjustment \cite{NEURIPS2023_973a0f50} technique. However, this approach might lead to difficulties and instability in optimization \cite{NEURIPS2019_621461af}. To tackle this issue, LfF \cite{NEURIPS2020_eddc3427} training a pair of neural networks simultaneously that can effectively mitigate the bias in the network. DebiasPL \cite{Wang_2022_CVPR} designs an adaptive debiasing module from a causal perspective. UniSSL \cite{huang2021universal} proposes class-sharing data detection and feature adaptation methods. DST \cite{NEURIPS2022_d10d6b28}  adopts a self-training debiased framework.  CDMAD \cite{Lee_2024_CVPR} uses solid color images to correct classifier bias during training and inference.  Based on this finding, we design the LCGC to train consistent gradients conflicting guided bias model from class-imbalanced data utilizing FixMatch \cite{Sohn_2020_nips} and ReMixMatch \cite{Berthelot_2020_iclr}.
\subsection{Class-Imbalanced Semi-Supervised Learning}
\label{sec:cissl}
This part of the research primarily leverages a fusion of CIL and SSL techniques, which allows for the full utilization of unlabeled data during training, while also providing them with balanced pseudo-labels. By assuming that the class distribution of the unlabeled set is known and the same as the labeled dataset. CReST \cite{wei2021crest} uses unlabeled samples predicted as the minority classes more frequently than those predicted as the majority classes for interactive self-training. CoSSL \cite{Cai_2021_ICCV} and BaCon \cite{Feng_Xie_Fang_Lin_2024} use an auxiliary classifier and train the classifier to be balanced. SMCLP \cite{DU2024110358} utilizes a collaborative manner to exploit the correlations from labels and instances to overcome the imbalanced problem. Conversely, SAW \cite{pmlr-v162-lai22b} mitigates class imbalance using smoothed reweighting based on the number of pseudo-labels belonging to each class without prior knowledge of the class distribution of the unlabeled set.

\section{Problem Setup and Preliminaries}
\label{sec:2}
\subsection{Problem Setup}
\label{sec:problem}
The problem of class-imbalanced semi-supervised learning aims to train a classifier involving labeled set $\mathcal{X}=\{(x^n, y^n):n \in (1, \cdots, N)\}$ and unlabeled set $\mathcal{U}=\{(u^m):m \in (1, \cdots, M)\}$, where $x^n \in \mathbb{R}^d$ and $y^n \in [C] =\{1, \cdots, C\}$ denote the $n$-th labeled sample and corresponding label, respectively, and $u^m \in \mathbb{R}^d$ denotes the $m$-th unlabeled sample. Here, the number of labeled and unlabeled samples of class $c$ as $N_c$ and $M_c$ i.e., $\sum^C_{c=1}N_c=N$ and $\sum^C_{c=1}M_c=M$, where $M_c$ is challenging to know in a realistic scenario.
Formally, the ratio of the class imbalance of labeled and unlabeled sets can be represented as $\gamma_l = \frac{N_1}{N_c} \geq 1 $ and $\gamma_u = \frac{M_1}{M_c} \geq 1$ under the class-imbalanced training set, where $N_1$ and $M_1$ are the number of samples in the largest labeled and unlabeled class. For each iteration of training, we sample minibatches $\mathcal{M}\mathcal{X} = \{(x^m_b, y^m_b): b \in (1, \cdots, B)\} \subset \mathcal{X}$ and $\mathcal{M}\mathcal{U} = \{(u^m_b):b \in (1, \cdots, \mu B)\} \subset \mathcal{U}$ from the training set, where $B$ denotes the minibatch size and $\mu$ denotes the relative size of $\mathcal{M}\mathcal{U}$ to $\mathcal{M}\mathcal{X}$. The goal of a CISSL model is to learn a classifier $f_\theta : \mathbb{R}^d \rightarrow \{1, \cdots, C\}$ that effectively classifies samples in a test set $\mathcal{X}^{test} = \{(x^{test}_k, y^{test}_k) \}$, where $\theta$ denotes parameters of base SSL algorithm. The output logits of $f_\theta$ on an input as $g_\theta(\cdot) \in \mathbb{R}^C$, i.e., $f_\theta(\cdot)=\arg\max_{c}g_\theta(\cdot)_c$, where $(\cdot)_c$ denotes the $c$-th element.
 
\subsection{Backbone SSL Algorithm}
\label{sec:backbone}
In this paper, we consider exploiting FixMatch \cite{Sohn_2020_nips} or ReMixMatch \cite{Berthelot_2020_iclr} as the backbone for the base CISSL problem. FixMatch first predicts the class probability of weakly augmented unlabeled data $\alpha(u^m_b)$ as $q_b = \text{Softmax} (g_\theta(\alpha(u^m_b))$ and then generates hard pseudo-label $\tilde{q}_b =\arg \max_c {q_{b,c}}$. Then the consistency regularization is calculated from $\tilde{q}_b$ and its strongly augmented $\mathcal{A}(u^m_b)$ version. ReMixMatch similarly produces $q_b$ but additionally adopts distribution alignment and sharpening strategies, which also conducts MixUp regularization and is self-supervised by rotating unlabeled samples \cite{gidaris_2018_iclr}. The training losses of FixMatch $\mathcal{L}_F$ and ReMixMatch $\mathcal{L}_R$ on $\mathcal{M}\mathcal{X}$ and $\mathcal{M}\mathcal{U}$ is given by:
\begin{equation}
    \mathcal{L}_F(\mathcal{M}\mathcal{X}, \mathcal{M}\mathcal{U}, \tilde{q}, \tau; \theta)
\end{equation}
\begin{equation}
    \mathcal{L}_R(\mathcal{M}\mathcal{X}, \mathcal{M}\mathcal{U}, \bar{q}; \theta)
\end{equation}
where $\tau$ denotes a predefined confidence threshold to improve the quality of the pseudo-labels. $\bar{q}$ denotes the sharpened pseudo-label after distribution alignments. Note that, in the absence of specifying FixMatch or ReMixMatch as the backbone for CISSL, we slightly abuse notation by uniformly using $\mathcal{L}$ to denote the training loss.

\subsection{Class-Distribution-Mismatch-Aware Debiasing} 
We primarily explore the reason why class-distribution-mismatch-aware debiasing (CDMAD) is effective \cite{Lee_2024_CVPR}.  CDMAD first calculates the logits on a weakly augmented unlabeled sample $g_\theta(\alpha(u^m_b))$, and the logits on a baseline (solid color image) $g_\theta(\mathcal{I})$. Then CDMAD adjusts for the classifier's biased degree by simple subtraction of both the training and testing phase.
\begin{equation}
    \textbf{\textit{Trainning phase: }} g^*_\theta = g_\theta(\alpha(u^m_b)) - g_\theta(\mathcal{I})
\end{equation}

\begin{equation}
    \textbf{\textit{Testing phase: }} g^*_\theta = g_\theta(x_k^{test}) - g_\theta(\mathcal{I})
\end{equation}
where $x_k^{logits}$ is test samples. 

\subsection{Integrated Gradients}
\label{sec:ig}
Focus on the image classification problem and assume the classifier network $f_{\theta}: \mathbb{R}^d \rightarrow \{1, \cdots, C\}$ is differentiable almost everywhere and perfectly fits the dataset, i.e., there exist interpolating parameters $\theta^*$ such that $\forall x_n \in \mathcal{X}: <x_n, \theta^*>=y_n$. Let $x_n \in \mathbb{R}^d$ be the input, and $\mathcal{I} \in \mathbb{R}^d$ be the baseline image, then the integrated gradients are obtained by accumulating the straightline path from the $\mathcal{I}$ to the $x_n$ \cite{pmlr_2017_icml}.
\begin{equation}
    \sum^d_{i=1} \text{IntergratedGrads}_i(x_n) = f_{\theta}(x_n) - f_{\theta}(\mathcal{I})
\end{equation}
\section{Why and How a Baseline Image Can Debias the Classifier Logits}
\label{sec:Why}

In this section, we characterize the class bias of pseudo-label $q_b$, which is generated by FixMatch or ReMixMatch from $g_\theta$. Note that the base SSL model parameters are updated according to the gradient flow:

\begin{equation}
    \frac{d\theta}{dt} = - \nabla_{\theta} \mathcal{L}(\mathcal{M}\mathcal{X}, \mathcal{M}\mathcal{U}, q; \theta)
\end{equation}

Previous work \cite{Lee_2024_CVPR} showed that the classifier's biased degree can be adjusted by a baseline image $\mathcal{I}$. We examine \textit{\textbf{why and how a baseline image can debias the classifier logits}}? 
We revisit an empirical phenomenon observed in CDMAD Section 4.4: After exploring a series of measures of baseline image debiasing, they find that the base CISSL framework can also be greatly improved by refinement of biased class predictions only during testing. As severe class distribution mismatch between labeled and unlabeled sets, the resulting classifier’s different biased degree toward each class. That is the logit distribution $P(g_\theta(\mathcal{I}))$ of a baseline image $\mathcal{I}$ through CISSL is imbalanced. We formulate the phenomena in the following assumption.

\begin{ass} The SSL classifiers trained on the class-imbalanced datasets produced biased results:
\begin{equation}
    P(y_n = c | x_n) \varpropto \exp{(g_{\theta,c}(x_n)))} 
\end{equation}
where $P(y_n = c | x_n) \neq 1/C$ for any $y_n \in [C]$. 
\end{ass}

For convenience, we define
\begin{equation}
    g’_\theta(x_n) = g_\theta(x_n) - g_\theta(\mathcal{I})
\end{equation}
where $g_\theta(x_n)$ is the logit of $x_n$. 

We then derive the conditional probability $P_b = P(y|g_{\theta}(x_n))$ and $P_a =P(y_n|g’_\theta(x_n))$ w.r.t. the outputs of $g_{\theta}(x_n)$ and $g’_{\theta}(x_n)$:

\begin{lem} For a biased classifier trained on the class-imbalanced datasets, the basic CISSL model's logit $g_{\theta}(x_n)$ and its refinement $g’_\theta(x_n)$ have diverse predictions:
\begin{equation}
    (g_{\theta}(x_n) \perp g’_\theta(x_n)) | y_n
\end{equation}
\end{lem}

We now state the main theorem regarding the class debias of the black baseline image below:
\begin{figure}
    \centering
    \includegraphics[width=1.0\columnwidth]{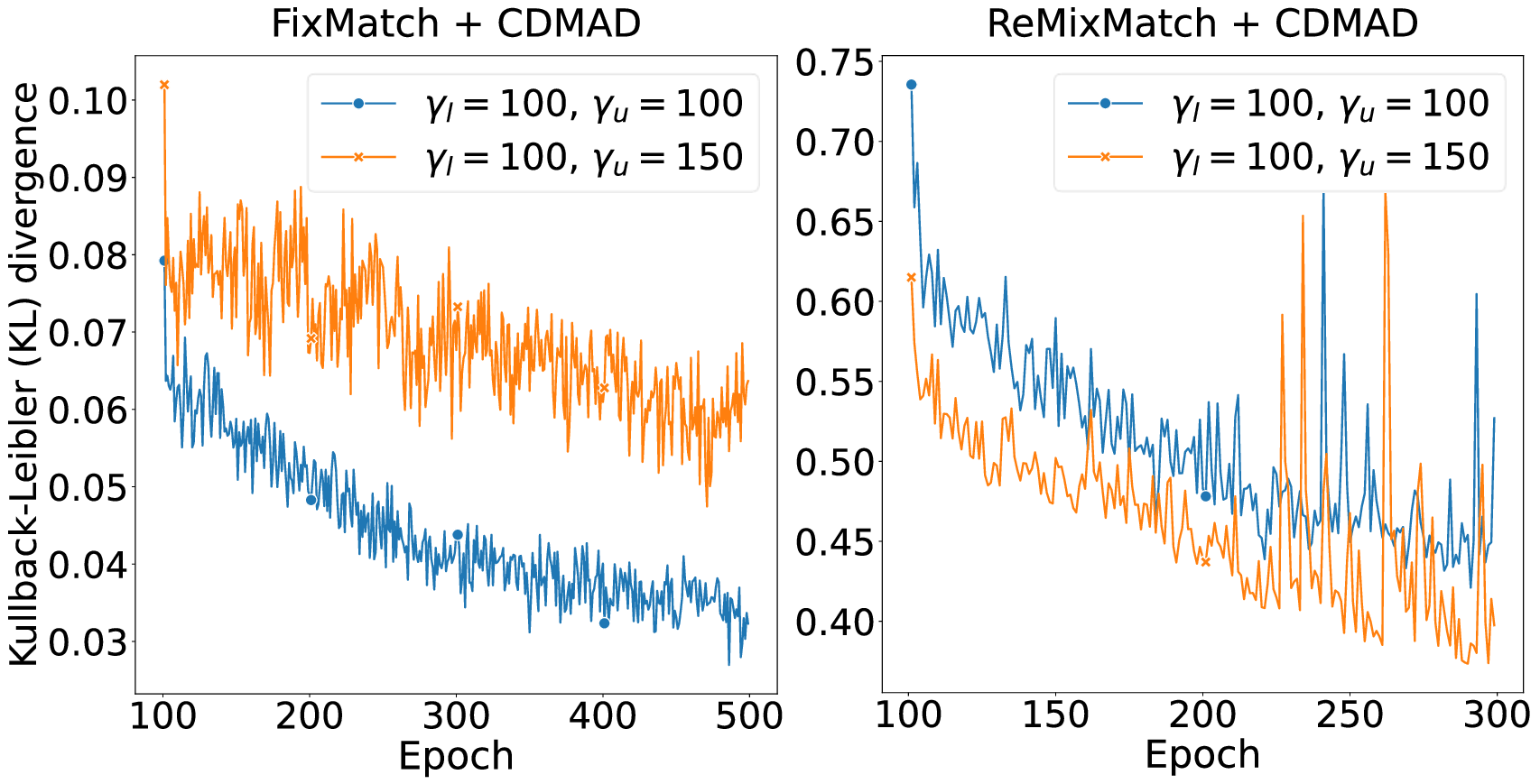}
    \caption{The plot of $\mathcal{L}_{kl}(\theta)$ under CIFAR10-LT dataset and different imbalance ratios $\gamma_l$, $\gamma_u$. Both columns show that refined by the baseline image, the KL divergence value continues to decrease, indicating that the logit output distribution of the model before and after refinement becomes closer.}
    \label{fig:kl_loss}
\end{figure}

\begin{thm} [Integrated gradient flow for class debiasing] For a CISSL classifier network $f_{\theta}$, if a baseline image (the baseline is preferably a black image) is exploited to refine the logit, i.e. $g’_\theta(\alpha(u^m_b)) =  g_\theta(\alpha(u^m_b)) - g_\theta(\mathcal{I})$ during the training phase, the gradient flow $- \nabla_{\theta} \mathcal{L}(\mathcal{M}\mathcal{X}, \mathcal{M}\mathcal{U}, q; \theta)$ of $f_{\theta}$ contains a linear integrated gradients term $\mathbb{I}(\theta)$:
\end{thm}
\begin{equation}
\begin{gathered}
    \nabla_{\theta} \mathcal{L}  = \mathbb{C}(\theta) + \mathbb{I}(\theta)\\
    \mathbb{I}(\theta) = - \sum_b (\sum^d_{i=1} \text{IntergratedGrads}_i(u_b^m)) \nabla g_{b}
\end{gathered}
\end{equation}
where $\mathbb{C}(\theta)$ is the term in the gradient flow other than the $\mathbb{I}(\theta)$ component.
\label{sec:LCGC}
\begin{figure*}[!htbp]
    \centering
    \includegraphics[width=1.8\columnwidth]{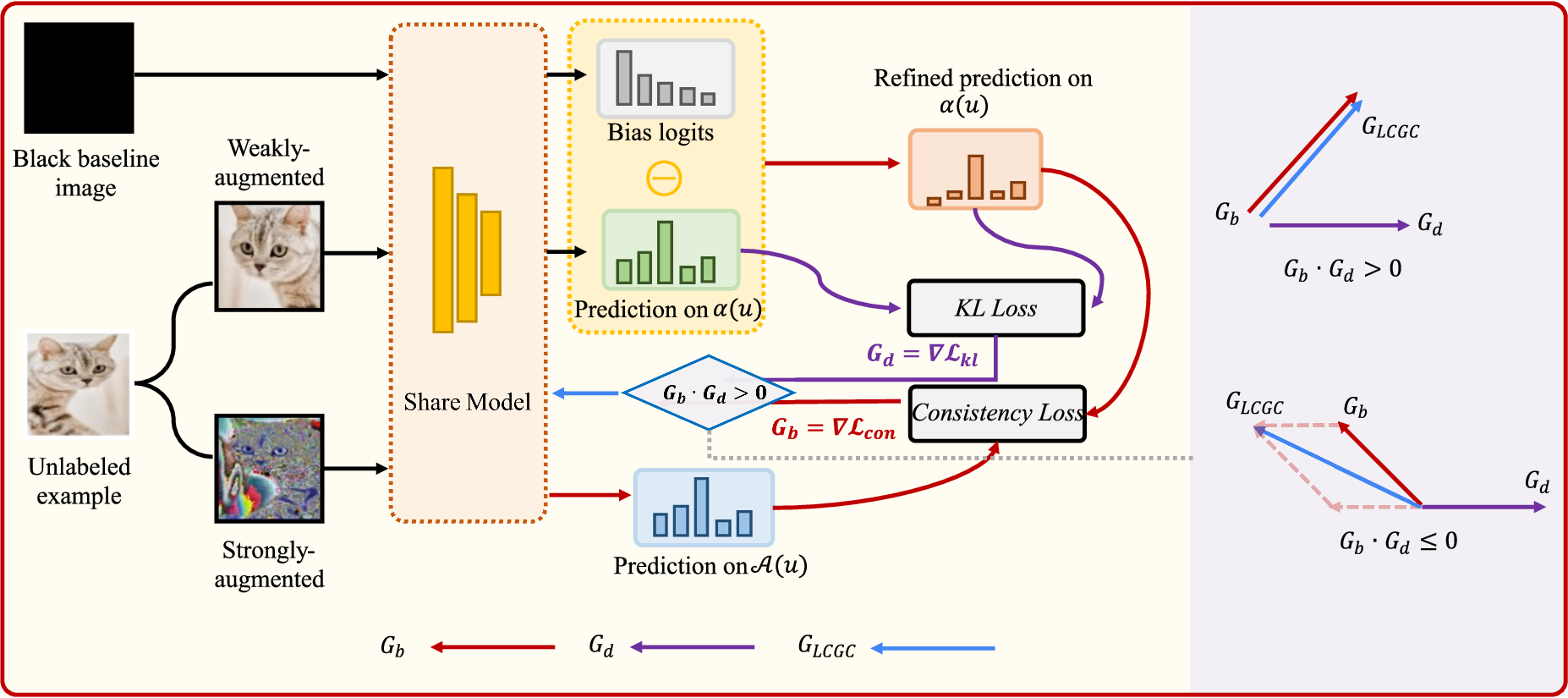}
    \caption{Pseudo-label refinement process using LCGC.}
    \label{fig:LCGC_process}
\end{figure*}
To derive the results, we first define some notations. Let $\mathcal{L}(\mathcal{M}\mathcal{X}, \mathcal{M}\mathcal{U}, q; \theta)= \mathcal{L}_{Con}(\mathcal{M}\mathcal{U}, \tilde{q}; \theta) + \mathcal{L}_{Sup}(\mathcal{M}\mathcal{X}; \theta)$ denote the loss function of backbone SSL algorithm on a minibatch for labeled set $\mathcal{M}\mathcal{X}$ and a minibatch for unlabeled set $\mathcal{M}\mathcal{U}$. Therefore, we have 
\begin{equation}
    \nabla_{\theta} \mathcal{L} = \nabla_{\theta} \mathcal{L}_{Con} + \nabla_{\theta} \mathcal{L}_{Sup}
\end{equation}
Since the adjust of CDMAD only acts on the $\mathcal{L}_{Con}$ term in $\mathcal{L}$, we only need to study $\nabla_{\theta} \mathcal{L}_{Con}$. Consequently, we have $\mathcal{L}_{Con}= \frac{1}{\mu B}\sum_{u^m_b \in \mathcal{M}\mathcal{U}} \mathbf{H}(P_{\theta} (y |\mathcal{A}(u^m_b)), \tilde{q}_b)$, where $\mathbf{H}$ is cross-entropy loss. It can be derived

\begin{equation}
	\begin{tiny}
		\begin{aligned}
			\nabla_{\theta} \mathcal{L}_{Con} &=  \sum_{b} \frac{\partial \mathcal{H}}{\partial q_{\mathcal{A}_b}} \frac{\partial q_{\mathcal{A}_b}}{\partial\theta} \\ 
			&= \sum_b q_{\mathcal{A}_b} \frac{\partial q_{\mathcal{A}_b}}{\partial\theta} - \sum_b \tilde{q}_b \frac{\partial q_{\mathcal{A}_b}}{\partial\theta} \\ 
			&= \sum_b q_{\mathcal{A}_b} \frac{\partial q_{\mathcal{A}_b}}{\partial\theta} - \sum_b (q_b - q_{\mathcal{I}}) \frac{\partial q_{\mathcal{A}_b}}{\partial\theta} \\ 
			&= \sum_b q_{\mathcal{A}_b} \frac{\partial q_{\mathcal{A}_b}}{\partial\theta} - \sum_b (g(u_b^m)_{b} - g(\mathcal{I})_{b}) \nabla g_b \\ 
			&= - \sum_b \left(\sum^d_{i=1} \text{IntergratedGrads}_i(u_b^m)\right) \nabla g_b + \sum_b q_{\mathcal{A}_b} \frac{\partial q_{\mathcal{A}_b}}{\partial\theta}
		\end{aligned}
	\end{tiny}
\end{equation}


Theorem 1 reveals during the training phase, the gradient flow is implicitly guided in the direction of the integrated gradient. With the attribute capability of integral gradient, the class-related features received more attention, and then the bias of the classifier was also reduced. Naturally, the best choice for the baseline image is the same as the baseline for the integrated gradient, which is \textbf{\textit{black}}.

Furthermore, we found an interesting phenomenon that the distributions of pseudo labels before and after refinement tend to be consistent, as the training process deepens. We employ KL divergence to measure the consistency of this distribution, as shown in Figure \ref{fig:kl_loss}. This observation leads us to the following corollary:

\begin{corollary} Under conditions of Theorem 1, if a black image is exploited for the refinement of pseudo-labels during training and if the gradient flow solution $\theta(\infty)$ satisfies that $f_{\theta(\infty)}(x_b) = y_b$, then gradient flow converges to a minimizer of the Kullback-Leibler (KL) divergence $\mathcal{L}_{kl}(\theta)$ between refinement and original pseudo-labels logits.
\begin{equation}
    \theta(\infty) = \arg \min_\theta \mathcal{L}_{kl}(\theta),\text{ }s.t.\text{ }f_{\theta(\infty)}(x_b) = y_b
\end{equation}

where 
\begin{equation}
    \footnotesize{
    \mathcal{L}_{kl} = -\sum_b g_\theta(x_b, \alpha(u^m_b)) \log \frac{g’_\theta(x_b, \alpha(u^m_b))}{g_\theta(x_b, \alpha(u^m_b))}}
    \label{Eq11}
\end{equation}
\end{corollary}

We now summarize the internal mechanism of the classifier debias with a black baseline image under the CISSL framework as follows. \textbf{1)} According to \citeauthor{pmlr_2017_icml}, for a pre-trained classifier network $f_{\theta}$, integrated gradients are better at reflecting class-related features of the input image. Since the CDMAD is implicitly trained in the direction of the linear integrated gradients, more label-irrelevant features can be discarded to achieve better classification results on SSL. \textbf{2)} According to Corollary 1.1, a cross-entropy minimization predictor in which the refinement process implicitly regulates the KL divergence between the logits. Therefore, as the KL divergence becomes smaller, the consistency of logits before and after debias increases, which further regularizes the model space and improves the generalization ability of a model.

\section{LCGC: A Consistency Gradient Conflicting Guided Debiasing Method}

From the analysis above, we further design a novel debiasing method that can improve the performance of the classifier under CISSL. In particular, our algorithm trains a classifier neural network $f_\theta$ as follows: 1) intentionally training the model $f_\theta$ to be \textbf{\textit{over-biased}} on the unlabeled training samples that exhibit inconsistency before and after logit debiasing and 2) using a black baseline image to refine of biased class predictions during testing. Figure \ref{fig:LCGC_process} illustrates the pseudolabel refinement process.
\subsection{Training a Consistency Conflicting Guided Model}
\label{sec:train}
Since we hope that the classifier can produce over-biased results, we need to make the distribution of pseudo-labels before and after refinement more different. Motivated by the success of regularization dropout \cite{NEURIPS2021_5a66b920}, we compare the debias logits $g'_\theta$ with the bias logits $g_\theta$ to regularize the gradient direction. Specifically, we obtain the \textit{\textbf{biased consistency direction}} $G_b$ by calculating the consistency regularization loss $\mathcal{L}_{Con}$ between the weakly augmented prediction $p(\alpha(u^m_b))$ and the strongly augmented prediction $p(\mathcal{A}(u^m_b))$ as follows:

\begin{equation}
    \mathcal{L}_{Con} = -\sum_b p(\mathcal{A}(u^m_b))\log{p(\alpha(u^m_b))}
\end{equation}
the \textit{\textbf{debiased direction}} $G_d$ based on the KL divergence according to Eq. (\ref{Eq11}).

We denote the gradients of $\mathcal{L}_{Con}$ and $\mathcal{L}_{kl}$ as $G_b = \nabla_{u} \mathcal{L}_{Con}(u)$ and $G_d = \nabla_{u} \mathcal{L}_{kl}(u)$, respectively. The relations between $G_b$ and $G_d$ are two-fold. \textbf{1)} Their angle is smaller than $90^\circ$, which indicates that the optimization direction of debias logits does not conflict with bias logits. In this case, we safely set the updated gradient direction $G_{LCGC}$ as $G_b$. \textbf{2)} Their angle is larger than $90^\circ$, which indicates that debias logits conflict with bias logits. That is, optimizing the classifier following $G_b$ will lead to an \textbf{\textit{over-biased}} classifier. In this case, we project the $G_b$ to the horizontal direction of $G_d$ then add it to $G_b$ to obtain an over-biased direction to optimize the classification model. Although this will slightly increase the KL loss, the post-adjustment logits by the black baseline image will be enhanced. Our LCGC strategy is mathematically formulated as:
\begin{equation}
G_{LCGC}=\begin{cases}
    G_b & \text{if } G_d \cdot G_b \geq 0 \\
    G_b + \lambda \cdot \frac{G_d \cdot G_b}{\|G_d\|^2}G_d & \text{otherwise}
\end{cases}
\label{grad}
\end{equation}
Instead of updating the parameter $\theta$ using $G_b$, we optimize the $\theta$ using $G_{LCGC}$, which encourages the gradient direction from biased classifier. We further introduce $\lambda$ in Eq. (\ref{grad}) to generalize the formulation, which can flexibly control the strength of debias logits guidance in applications. In particular, $\lambda=1$ denotes projecting $G_b$ to the \textbf{\textit{over-biased}} direction, while setting $\lambda=0$ makes LCGC degenerate to CDMAD, \textit{i.e.}, CDMAD is a special case of our strategy. See more details on the selection of parameter $\lambda$ in Section {5.3}.

\subsection{Refinement of Biased Logits During Inference}  
\label{ssec:inference}
While we train a biased model as described earlier, we must refine biased logit during inference on $x^{test}_k$, for $k=1, \cdots, K$. Then, the logits for test samples:
\begin{equation}
    g^*_\theta(x^{test}_k) = g'_\theta(x^{test}_k) -g_\theta(\mathcal{I})
\end{equation}
where $g^*_\theta(x^{test}_k)$ is the finally prediction of $x^{test}_k$.

\section{Experiments}
\label{sec:experiment}
\subsection{Experimental Setup}
We implemented a series of experiments on CIFAR-10-LT, CIFAR-100-LT \cite{Cui_2019_CVPR}, SVHN-LT \cite{miyato2018virtual} and STL-100-LT \cite{Kim_2020_nips} following the settings of previous studies \cite{Fan_2022_CVPR, Lee_2024_CVPR}. To assess classification performance across these datasets, we employed balanced accuracy (bACC) \cite{Huang_2016_CVPR} and geometric mean (GM) \cite{Kubat_1997_Icml} as metrics for CIFAR-10-LT, SVHN-LT and STL-10-LT. For CIFAR-100-LT, evaluation was conducted using only the bACC metric. Each experiment was conducted three times on RTX 3090 GPUs to ensure reproducibility, and we report both the average and standard error of the performance measures.
\subsection{Experimental Results}
\label{sec:exre}
Table \ref{Tab:cifar10_known} presents a detailed comparison of bACC and GM across various algorithms on CIFAR-10-LT under conditions where $\gamma_u$ is known and matched to $\gamma_l$. It is observed that SSL algorithms like FixMatch and ReMixMatch demonstrate marked improvements in classification performance over the traditional approach. Despite their improved performance, they fall short of the efficiencies achieved by CISSL algorithms, emphasizing the necessity of directly addressing class imbalance issues within SSL frameworks. Notably, CISSL algorithms integrating the LCGC approach consistently outperform other methodologies. This strategy markedly improves performance across varying class imbalance ratios, as evidenced by superior bACC and GM scores.

\begin{table}[!b]
  \centering
  \scriptsize
  \setlength{\tabcolsep}{1mm}
    \begin{tabular}{cccc}
    \toprule
    \multicolumn{4}{c}{CIFAR-10-LT ($\gamma = \gamma_l = \gamma_u$, $\gamma_u$ is assumed to be known)}\\
    \midrule
    Algorithm &   $\gamma=50$   &  $\gamma=100$      & $\gamma=150$ \\
    \midrule
    Vanilla & \numstd{65.2}{0.05}/\numstd{61.1}{0.09} & \numstd{58.8}{0.13}/\numstd{58.2}{0.11} & \numstd{55.6}{0.43}/\numstd{44.0}{0.98} \\
    \midrule
    Re-sampling & \numstd{64.3}{0.48}/\numstd{60.6}{0.67} & \numstd{55.8}{0.47}/\numstd{45.1}{0.30} & \numstd{52.2}{0.05}/\numstd{38.2}{1.49}\\
    LDAM-DRW & \numstd{68.9}{0.07}/\numstd{67.0}{0.08} & \numstd{62.8}{0.17}/\numstd{58.9}{0.60} & \numstd{57.9}{0.20}/\numstd{50.4}{0.30}\\
    cRT   & \numstd{67.8}{0.13}/\numstd{66.3}{0.15} & \numstd{63.2}{0.45}/\numstd{59.9}{0.40} & \numstd{59.3}{0.10}/\numstd{54.6}{0.72}\\
    \midrule
    FixMatch &\multicolumn{1}{c}{\numstd{79.2}{0.33}/\numstd{77.8}{0.36}} & \multicolumn{1}{c}{\numstd{71.5}{0.72}/\numstd{66.8}{1.51}} & \multicolumn{1}{c}{\numstd{68.4}{0.15}/\numstd{59.9}{0.43}} \\
    /+DARP+cRT & \numstd{85.8}{0.43}/\numstd{85.6}{0.56} & \numstd{82.4}{0.26}/\numstd{81.8}{0.17} & \numstd{79.6}{0.42}/\numstd{78.9}{0.35} \\
    /+CReST+LA & \multicolumn{1}{c}{\numstd{85.6}{0.36}/\numstd{81.9}{0.45}} & \multicolumn{1}{c}{\numstd{81.2}{0.70}/\numstd{74.5}{0.99}} & \multicolumn{1}{c}{\numstd{71.9}{2.24}/\numstd{64.4}{1.75}} \\
    /+ABC & \multicolumn{1}{c}{\numstd{85.6}{0.26}/\numstd{85.2}{0.29}} & \multicolumn{1}{c}{\numstd{81.1}{1.14}/\numstd{80.3}{1.29}} & \multicolumn{1}{c}{\numstd{77.3}{1.25}/\numstd{75.6}{1.65}} \\
    /+CoSSL & \multicolumn{1}{c}{\numstd{86.8}{0.30}/\numstd{86.6}{0.25}} & \multicolumn{1}{c}{\numstd{83.2}{0.49}/\numstd{82.7}{0.60}} & \multicolumn{1}{c}{\numstd{80.3}{0.55}/\numstd{79.6}{0.57}} \\
    /+SAW+LA & \multicolumn{1}{c}{\numstd{86.2}{0.15}/\numstd{83.9}{0.35}} & \multicolumn{1}{c}{\numstd{80.7}{0.15}/\numstd{77.5}{0.21}} & \multicolumn{1}{c}{\numstd{73.7}{0.06}/\numstd{71.2}{0.17}} \\
    /+Adsh & \multicolumn{1}{c}{\numstd{83.4}{0.06}/ -} & \multicolumn{1}{c}{\numstd{76.5}{0.35}/ -} & \multicolumn{1}{c}{\numstd{71.5}{0.30}/ -}\\
    /+DebiasPL & \multicolumn{1}{c}{- / -} & \multicolumn{1}{c}{\numstd{80.6}{0.50}/ -} & \multicolumn{1}{c}{- / -} \\
    /+UDAL & \multicolumn{1}{c}{\numstd{86.5}{0.29}/ -} & \multicolumn{1}{c}{\numstd{81.4}{0.39}/ -} & \multicolumn{1}{c}{\numstd{77.9}{0.33}/ -}\\
    /+L2AC & \multicolumn{1}{c}{- / -} & \multicolumn{1}{c}{\numstd{82.1}{0.57}/\numstd{81.5}{0.64}} & \multicolumn{1}{c}{\numstd{77.6}{0.53}/\numstd{75.8}{0.71}} \\
    /+CDMAD & \multicolumn{1}{c}{\numstd{87.3}{0.12}/\numstd{87.0}{0.15}} & \multicolumn{1}{c}{\numstd{83.6}{0.46}/\numstd{83.1}{0.57}} & \multicolumn{1}{c}{\numstd{80.8}{0.86}/\numstd{79.9}{1.07}}\\
    \rowcolor[rgb]{ .949,  .949,  .949} /+LCGC & \multicolumn{1}{c}{\numstd{\textbf{87.3}}{0.03}/\numstd{\textbf{87.1}}{0.07}} & \multicolumn{1}{c}{\numstd{\textbf{84.9}}{0.13}/\numstd{\textbf{84.6}}{0.17}} & \multicolumn{1}{c}{\numstd{\textbf{82.4}}{0.01}/\numstd{\textbf{81.9}}{0.09}} \\
    \midrule
    ReMixMatch &  \multicolumn{1}{c}{\numstd{81.5}{0.26}/\numstd{80.2}{0.32}} & \multicolumn{1}{c}{\numstd{73.8}{0.38}/\numstd{69.5}{0.84}} & \multicolumn{1}{c}{\numstd{69.9}{0.47}/\numstd{62.5}{0.35}} \\
    /+DARP+cRT &  \multicolumn{1}{c}{\numstd{87.3}{0.61}/\numstd{87.0}{0.11}} & \multicolumn{1}{c}{\numstd{83.5}{0.07}/\numstd{83.1}{0.09}} & \multicolumn{1}{c}{\numstd{79.7}{0.54}/\numstd{78.9}{0.49}} \\
    /+CReST+LA &  \multicolumn{1}{c}{\numstd{84.2}{0.11}/ -} & \multicolumn{1}{c}{\numstd{81.3}{0.34}/ -} & \multicolumn{1}{c}{\numstd{79.2}{0.31}/ -} \\
    /+ABC &  \multicolumn{1}{c}{\numstd{87.9}{0.47}/\numstd{87.6}{0.51}} & \multicolumn{1}{c}{\numstd{84.5}{0.32}/\numstd{84.1}{0.36}} & \multicolumn{1}{c}{\numstd{80.5}{1.18}/\numstd{79.5}{1.36}} \\
    /+CoSSL &   \multicolumn{1}{c}{\numstd{87.7}{0.21}/\numstd{87.6}{0.25}} & \multicolumn{1}{c}{\numstd{84.1}{0.56}/\numstd{83.7}{0.66}} & \multicolumn{1}{c}{\numstd{81.3}{0.83}/\numstd{80.5}{0.76}} \\
    /+SAW+cRT &  \multicolumn{1}{c}{\numstd{87.6}{0.21}/\numstd{87.4}{0.26}} & \multicolumn{1}{c}{\numstd{85.4}{0.32}/\numstd{83.9}{0.21}} & \multicolumn{1}{c}{\numstd{79.9}{0.15}/\numstd{79.9}{0.12}} \\
    /+CDMAD &  \multicolumn{1}{c}{\numstd{88.3}{0.35}/\numstd{88.1}{0.35}} & \multicolumn{1}{c}{\numstd{85.5}{0.46}/\numstd{85.3}{0.44}} & \multicolumn{1}{c}{\numstd{82.5}{0.23}/\numstd{82.0}{0.30}} \\
    \rowcolor[rgb]{ .949,  .949,  .949} /+LCGC &  \multicolumn{1}{c}{\numstd{\textbf{88.7}}{0.14}/\numstd{\textbf{88.5}}{0.14}} & \multicolumn{1}{c}{\numstd{\textbf{85.7}}{0.42}/\numstd{\textbf{85.4}}{0.44}} & \multicolumn{1}{c}{\numstd{\textbf{82.8}}{0.31}/\numstd{\textbf{82.4}}{0.37}}\\
    
    \bottomrule
    \end{tabular}
 \caption{Comparison of bACC/GM on CIFAR-10-LT under $\gamma = \gamma_l = \gamma_u$ ($\gamma_u$ is assumed to be known).}
\label{Tab:cifar10_known}
\end{table}

Table \ref{Tab:cifar10_unknown} presents a comparative analysis of bACC and GM metrics on CIFAR-10-LT datasets under conditions where the class distribution of labeled and unlabeled sets are mismatched ($\gamma_l \neq \gamma_u$). The results highlight the robustness and efficacy of the LCGC approach in addressing class distribution mismatches in semi-supervised learning scenarios. For the CIFAR-10-LT dataset, where $\gamma_l$ is fixed at 100 and $\gamma_u$ varies, the performance of FixMatch combined with LCGC is notably superior. We outperform the second-best by an absolute accuracy of 1.7$\%$ at an imbalance ratio $\gamma_u = 150$ with FixMatch, showcasing the ability of LCGC to mitigate biases introduced by class distribution mismatches effectively. Note that we employed ReMixMatch* \cite{NEURIPS2020_a7968b43} to improve baseline algorithms, significantly improving the classification performance. However, LCGC still achieved better results in real-world scenarios. This makes LCGC more effective and scalable where the class distribution of unlabeled data is unknown and difficult to estimate. 

\begin{figure*}[!htbp]
    \centering
    \includegraphics[width=2.1\columnwidth]{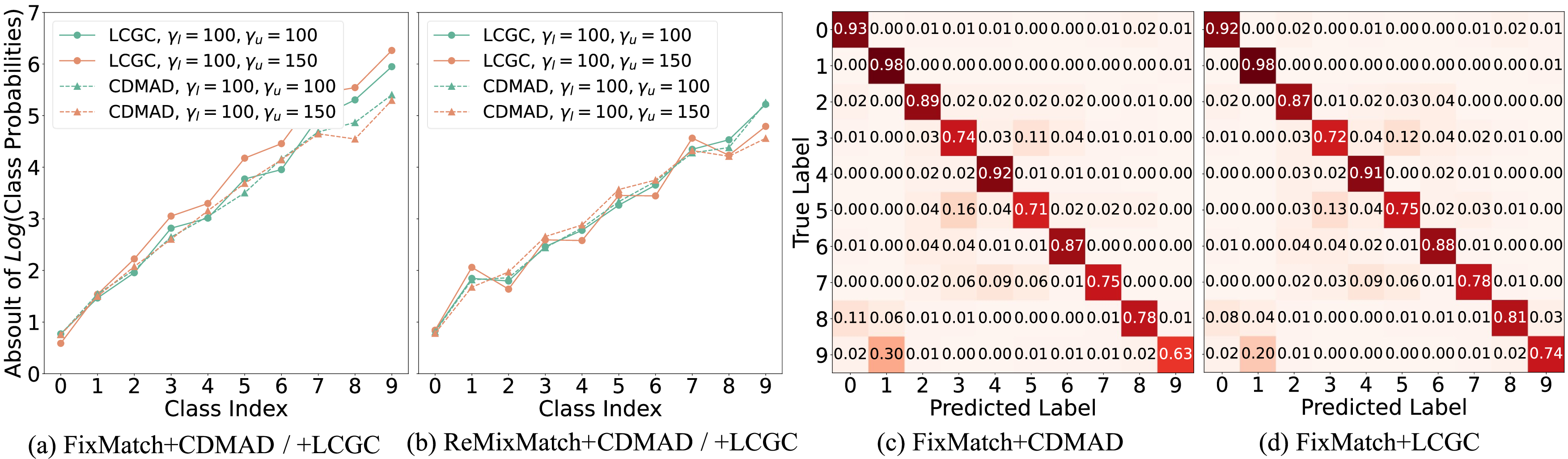}
    \caption{(a) and (b) present the class probabilities that take the absolute value of the log predicted on a black image using the proposed algorithm. (c) and (d) present the confusion matrices of the class predictions on test samples.}
    \label{fig:confusion_matrices}
\end{figure*}

\begin{table}[!b]
  \centering
  \scriptsize
   \setlength{\tabcolsep}{1mm}
    \begin{tabular}{cccc}
    \toprule
    \multicolumn{4}{c}{CIFAR-10-LT($\gamma_l = 100$, $\gamma_u$ is assumed to be unknown)} \\
    \midrule
    Algorithm &   $\gamma_u=1$   & $\gamma_u=50$   &  $\gamma_u=150$\\
    \midrule
    FixMatch & \numstd{68.9}{1.95}/\numstd{42.8}{8.11} & \numstd{73.9}{0.25}/\numstd{70.5}{0.52} & \numstd{69.6}{0.60}/\numstd{62.6}{1.11}\\
    /+DARP & \numstd{85.4}{0.55}/\numstd{85.0}{0.65} & \numstd{77.3}{0.17}/\numstd{75.5}{0.21} & \numstd{72.9}{0.24}/\numstd{69.5}{0.18}\\
    /+DAPR+LA & \numstd{86.6}{1.11}/\numstd{86.2}{1.15} & \numstd{82.3}{0.32}/\numstd{81.5}{0.29} & \numstd{78.9}{0.23}/\numstd{77.7}{0.06}\\
    /+DARP+cRT & \numstd{87.0}{0.70}/\numstd{86.8}{0.67} & \numstd{82.7}{0.21}/\numstd{82.3}{0.25} & \numstd{80.7}{0.44}/\numstd{80.2}{0.61}\\
    /+ABC & \numstd{82.7}{0.49}/\numstd{81.9}{0.68} & \numstd{82.7}{0.64}/\numstd{82.0}{0.76} & \numstd{78.4}{0.87}/\numstd{77.2}{1.07} \\
    /+SAW & \numstd{81.2}{0.68}/\numstd{80.2}{0.91} & \numstd{79.8}{0.25}/\numstd{79.1}{0.32} & \numstd{74.5}{0.97}/\numstd{72.5}{1.37}\\
    /+SAW+LA & \numstd{84.5}{0.68}/\numstd{84.1}{0.26} & \numstd{82.9}{0.38}/\numstd{82.6}{0.38} & \numstd{79.1}{0.81}/\numstd{78.6}{0.91} \\
    /+SAW+cRT & \numstd{84.6}{0.23}/\numstd{84.4}{0.26} & \numstd{81.6}{0.38}/\numstd{81.3}{0.32} & \numstd{77.6}{0.40}/\numstd{77.1}{0.41} \\
    /+CDMAD & \numstd{87.5}{0.46}/\numstd{87.1}{0.50} & \numstd{85.7}{0.36}/\numstd{85.3}{0.38} & \numstd{82.3}{0.23}/\numstd{81.8}{0.29} \\
    \rowcolor[rgb]{ .949,  .949,  .949} /+LCGC & \numstd{\textbf{88.2}}{0.42}/\numstd{\textbf{87.8}}{0.44} & \numstd{\textbf{85.9}}{0.44}/\numstd{\textbf{85.4}}{0.35} & \numstd{\textbf{84.0}}{0.20}/\numstd{\textbf{83.7}}{0.19} \\
    \midrule
    ReMixMatch & \numstd{48.3}{0.14}/\numstd{19.5}{0.85} & \numstd{75.1}{0.43}/\numstd{71.9}{0.77} & \numstd{72.5}{0.10}/\numstd{68.2}{0.32} \\
    ReMixMatch* & \numstd{85.0}{1.35}/\numstd{84.3}{1.55} & \numstd{77.0}{0.12}/\numstd{74.7}{0.04} & \numstd{72.8}{0.10}/\numstd{68.8}{0.21} \\
    /+DARP & \numstd{86.9}{0.10}/\numstd{86.4}{0.15} & \numstd{77.4}{0.22}/\numstd{75.0}{0.25} & \numstd{73.2}{0.11}/\numstd{69.2}{0.31}\\
    /+DARP+LA & \numstd{81.8}{0.45}/\numstd{80.9}{0.40} & \numstd{83.9}{0.42}/\numstd{83.4}{0.45} & \numstd{81.1}{0.20}/\numstd{80.3}{0.26}\\
    /+DARP+cRT & \numstd{88.7}{0.25}/\numstd{88.5}{0.25} & \numstd{83.5}{0.53}/\numstd{83.1}{0.51} & \numstd{80.9}{0.25}/\numstd{80.3}{0.31} \\
    /+ABC & \numstd{76.4}{5.34}/\numstd{74.8}{6.05} & \numstd{85.2}{0.20}/\numstd{84.7}{0.25} & \numstd{80.4}{0.40}/\numstd{80.0}{0.44}\\
    /+SAW & \numstd{87.0}{0.75}/\numstd{86.4}{0.85} & \numstd{80.6}{1.57}/\numstd{79.2}{2.19} & \numstd{77.6}{0.76}/\numstd{76.0}{0.93}\\
    /+SAW+LA & \numstd{74.2}{1.49}/\numstd{65.1}{2.36} & \numstd{84.8}{1.07}/\numstd{82.4}{2.32} & \numstd{81.3}{2.42}/\numstd{80.9}{2.47}\\
    /+SAW+cRT & \numstd{88.8}{0.79}/\numstd{88.6}{0.83} & \numstd{84.5}{0.78}/\numstd{83.6}{1.27} & \numstd{82.4}{0.10}/\numstd{82.0}{0.10} \\
    /+CDMAD & \numstd{89.9}{0.46}/\numstd{89.6}{0.46} & \numstd{86.9}{0.21}/\numstd{86.7}{0.17} & \numstd{83.1}{0.46}/\numstd{82.7}{0.50} \\
    \rowcolor[rgb]{ .949,  .949,  .949} /+LCGC & \numstd{\textbf{90.1}}{0.48}/\numstd{\textbf{89.6}}{0.43} & \numstd{\textbf{87.0}}{0.13}/\numstd{\textbf{86.8}}{0.13} & \numstd{\textbf{83.9}}{0.31}/\numstd{\textbf{83.6}}{0.33}\\
    \bottomrule
    \end{tabular}%
  \caption{Comparison of bACC/GM on CIFAR-10-LT under $\gamma_l \neq \gamma_u$ ($\gamma_u$ is assumed to be unknown). ReMixMatch* denotes ReMixMatch with the estimated class distribution of the unlabeled set.}
  \label{Tab:cifar10_unknown}
  
\end{table}%

\begin{table}[!b]
  \centering
  \scriptsize
  \setlength{\tabcolsep}{1mm}
  \begin{tabular}{ccccccc}
  \toprule
  \multicolumn{7}{c}{CIFAR-10-LT, $\gamma_l = 100$, $\gamma_u = 100$ (reversed)} \\
  \midrule
    Algorithm &   ABC   &  SAW   & SAW+LA  &  SAW+cRT &  CDMAD & LCGC\\
    \midrule
    FixMatch+ &\multicolumn{1}{c}{69.5/66.8} & \multicolumn{1}{c}{72.3/68.7} & \multicolumn{1}{c}{74.1/72.0} & \multicolumn{1}{c}{75.5/73.9} & \multicolumn{1}{c}{77.1/75.4} & \multicolumn{1}{c}{\textbf{78.6}/\textbf{77.2}} \\
    ReMixMatch+ &\multicolumn{1}{c}{63.6/60.5} & \multicolumn{1}{c}{79.5/78.5} & \multicolumn{1}{c}{50.2/14.8} & \multicolumn{1}{c}{80.8/79.9} & \multicolumn{1}{c}{81.7/81.0} & \multicolumn{1}{c}{\textbf{83.6}/\textbf{82.4}}\\
  \bottomrule
  \end{tabular}%
\caption{Comparison of bACC/GM under $\gamma_l = \gamma_u =100$ (reversed)}
\label{Tab:reversed}
\end{table}%
To examine the performance of our method in more imbalanced scenarios, experiments were also conducted under conditions where the class distribution of the unlabeled set was imbalanced in the opposite direction to the labeled set. As shown in Table \ref{Tab:reversed}, LCGC consistently outperforms other algorithms under these settings.

Table \ref{Tab:cifar100_known} presents a detailed comparison of bACC on CIFAR-100-LT across various algorithms under different class imbalance ratios ($\gamma$). The results clearly demonstrate the effectiveness of the proposed LCGC approach in improving classification performance under class-imbalanced settings with a large number of classes. When comparing FixMatch and ReMixMatch combined with various debiasing techniques, it is evident that LCGC consistently achieves the highest bACC scores. In scenarios where $\gamma$ is set to 20, 50, and 100, FixMatch+LCGC and ReMixMatch+LCGC outperform other methods. Besides the best performance across settings, our method also improves performance for small imbalance ratios as well (1.0$\%$ higher than the second-best at imbalance ratio $\gamma = 20$ with FixMatch).

\begin{table}[!b]
  \centering
  \scriptsize
    \begin{tabular}{cccc}
    \toprule
    \multicolumn{4}{c}{CIFAR-100-LT($\gamma = \gamma_l = \gamma_u$)} \\
    \midrule
    Algorithm &   $\gamma=20$   &  $\gamma=50$      & $\gamma=100$  \\
    \midrule
    FixMatch &\multicolumn{1}{c}{\numstd{49.6}{0.78}} & \multicolumn{1}{c}{\numstd{42.1}{0.33}} & \multicolumn{1}{c}{\numstd{37.6}{0.48}}\\
    FixMatch+DARP &\multicolumn{1}{c}{\numstd{50.8}{0.77}} & \multicolumn{1}{c}{\numstd{43.1}{0.54}} & \multicolumn{1}{c}{\numstd{38.3}{0.47}}\\
    FixMatch+DARP+cRT &\multicolumn{1}{c}{\numstd{51.4}{0.68}} & \multicolumn{1}{c}{\numstd{44.9}{0.54}} & \multicolumn{1}{c}{\numstd{40.4}{0.78}}\\
    FixMatch+CReST &\multicolumn{1}{c}{\numstd{51.8}{0.68}} & \multicolumn{1}{c}{\numstd{44.9}{0.50}} & \multicolumn{1}{c}{\numstd{40.1}{0.65}}\\
    FixMatch+CReST+LA &\multicolumn{1}{c}{\numstd{52.9}{0.07}} & \multicolumn{1}{c}{\numstd{47.3}{0.17}} & \multicolumn{1}{c}{\numstd{42.7}{0.70}}\\
    FixMatch+ABC &\multicolumn{1}{c}{\numstd{53.3}{0.79}} & \multicolumn{1}{c}{\numstd{46.7}{0.26}} & \multicolumn{1}{c}{\numstd{41.2}{0.65}}\\
    FixMatch+CoSSL &\multicolumn{1}{c}{\numstd{53.9}{0.78}} & \multicolumn{1}{c}{\numstd{47.6}{0.57}} & \multicolumn{1}{c}{\numstd{43.0}{0.61}}\\
    FixMatch+UDAL &\multicolumn{1}{c}{-} & \multicolumn{1}{c}{\numstd{48.0}{0.56}} & \multicolumn{1}{c}{\numstd{43.7}{0.41}}\\
    FixMatch+CDMAD &\multicolumn{1}{c}{\numstd{54.3}{0.44}} & \multicolumn{1}{c}{\numstd{48.8}{0.75}} & \multicolumn{1}{c}{\numstd{44.1}{0.29}}\\
    \rowcolor[rgb]{ .949,  .949,  .949} FixMatch+LCGC &\multicolumn{1}{c}{\numstd{\textbf{55.3}}{0.48}} & \multicolumn{1}{c}{\numstd{\textbf{49.3}}{0.26}} & \multicolumn{1}{c}{\numstd{\textbf{44.8}}{0.47}}\\
    \midrule
    ReMixMatch &\multicolumn{1}{c}{\numstd{51.6}{0.43}} & \multicolumn{1}{c}{\numstd{44.2}{0.59}} & \multicolumn{1}{c}{\numstd{39.3}{0.43}}\\
    ReMixMatch+DARP &\multicolumn{1}{c}{\numstd{51.9}{0.35}} & \multicolumn{1}{c}{\numstd{44.7}{0.66}} & \multicolumn{1}{c}{\numstd{39.8}{0.53}}\\
    ReMixMatch+DARP+cRT &\multicolumn{1}{c}{\numstd{54.5}{0.42}} & \multicolumn{1}{c}{\numstd{48.5}{0.91}} & \multicolumn{1}{c}{\numstd{43.7}{0.81}}\\
    ReMixMatch+CReST &\multicolumn{1}{c}{\numstd{51.3}{0.34}} & \multicolumn{1}{c}{\numstd{45.5}{0.76}} & \multicolumn{1}{c}{\numstd{41.0}{0.78}}\\
    ReMixMatch+CReST+LA &\multicolumn{1}{c}{\numstd{51.9}{0.60}} & \multicolumn{1}{c}{\numstd{46.6}{1.14}} & \multicolumn{1}{c}{\numstd{41.7}{0.69}}\\
    ReMixMatch+ABC &\multicolumn{1}{c}{\numstd{55.6}{0.35}} & \multicolumn{1}{c}{\numstd{47.9}{0.10}} & \multicolumn{1}{c}{\numstd{42.2}{0.12}}\\
    ReMixMatch+CoSSL &\multicolumn{1}{c}{\numstd{55.8}{0.62}} & \multicolumn{1}{c}{\numstd{48.9}{0.61}} & \multicolumn{1}{c}{\numstd{44.1}{0.59}}\\
    ReMixMatch+CDMAD &\multicolumn{1}{c}{\numstd{57.0}{0.32}} & \multicolumn{1}{c}{\numstd{\textbf{51.1}}{0.46}} & \multicolumn{1}{c}{\numstd{44.9}{0.42}}\\
    \rowcolor[rgb]{ .949,  .949,  .949} ReMixMatch+LCGC &\multicolumn{1}{c}{\numstd{\textbf{57.3}}{0.32}} & \multicolumn{1}{c}{\numstd{50.7}{0.38}} & \multicolumn{1}{c}{\numstd{\textbf{45.9}}{0.64}}\\
    \bottomrule
    \end{tabular}%
  \caption{Comparison of bACC on CIFAR-100-LT}
  \label{Tab:cifar100_known}
\end{table}%

\begin{table}[!b]
  \centering
  \scriptsize
  \setlength{\tabcolsep}{1mm}
    \begin{tabular}{cccc}
    \toprule
    \multicolumn{3}{c}{STL-10-LT} &\multicolumn{1}{c}{SVHN-LT}\\
    \midrule
    Algorithm &   $\gamma_l=10$  & $\gamma_l=20$ & $\gamma_l=100$ \\
    \midrule
    FixMatch &\multicolumn{1}{c}{\numstd{72.9}{0.09}/\numstd{69.6}{0.01}} & \multicolumn{1}{c}{\numstd{63.4}{0.21}/\numstd{52.6}{0.09}}& \multicolumn{1}{c}{\numstd{88.0}{0.30}/\numstd{79.4}{0.54}}\\
    /+DARP & \multicolumn{1}{c}{\numstd{77.8}{0.33}/\numstd{76.5}{0.40}} & \multicolumn{1}{c}{\numstd{69.9}{1.77}/\numstd{65.4}{3.07}}&\multicolumn{1}{c}{\numstd{88.6}{0.19}/\numstd{80.5}{0.54}}\\
    /+DAPR+LA & \multicolumn{1}{c}{\numstd{78.6}{0.30}/\numstd{77.4}{0.40}} & \multicolumn{1}{c}{\numstd{71.9}{0.49}/\numstd{68.7}{0.51}}& - / -\\
    /+DARP+cRT & \multicolumn{1}{c}{\numstd{79.3}{0.23}/\numstd{78.7}{0.21}} & \multicolumn{1}{c}{\numstd{74.1}{0.61}/\numstd{73.1}{1.21}}&\numstd{89.9}{0.44}/\numstd{83.5}{0.61}\\
    /+ABC & \multicolumn{1}{c}{\numstd{79.1}{0.46}/\numstd{78.1}{0.57}} & \multicolumn{1}{c}{\numstd{73.8}{0.15}/\numstd{72.1}{0.15}}&\numstd{92.0}{0.38}/\numstd{87.9}{0.73}\\
    /+CDMAD & \multicolumn{1}{c}{\numstd{79.9}{0.23}/\numstd{78.9}{0.38}} & \multicolumn{1}{c}{\numstd{75.2}{0.40}/\numstd{73.5}{0.31}}&\numstd{92.4}{0.16}/\numstd{92.2}{0.17}\\
    \rowcolor[rgb]{ .949,  .949,  .949} /+LCGC & \multicolumn{1}{c}{\numstd{\textbf{80.1}}{0.42}/\numstd{\textbf{79.2}}{0.27}}& \multicolumn{1}{c}{\numstd{\textbf{76.6}}{0.34}/\numstd{\textbf{75.2}}{0.34}}&\numstd{\textbf{93.3}}{0.16}/\numstd{\textbf{93.2}}{0.17}\\
    \bottomrule
    \end{tabular}%
  \caption{Comparison of bACC/GM on STL-10-LT and SVHN-LT ($\gamma_u$ is Unknown)}
  \label{Tab:STL_SVHN}
\end{table}%
Table \ref{Tab:STL_SVHN} summarizes bACC and GM of the baseline algorithms on STL-10-LT and SVHN-LT. In the STL-10-LT dataset, where $\gamma_l$ varies, FixMatch combined with LCGC again demonstrates superior performance. With $\gamma_l$ set to 10, FixMatch+LCGC achieves better results than other algorithms, and when $\gamma_l$ is increased to 20, the performance remains consistently higher. In the SVHN-LT dataset, LCGC outperformed the baseline algorithms by a large margin. The effective use of unlabeled samples through gradient-guided pseudo-label refinement may allow the proposed algorithm to be suitable for CISSL on special scenarios datasets.

\begin{table}[!b]
  \centering
    \begin{tabular}{ccc}
    \toprule
    \multicolumn{3}{c}{Small-ImageNet-127 ($\gamma= \gamma_l =\gamma_u$)} \\
    \midrule
    Algorithm &   32 $\times$ 32 & 64 $\times$ 64\\
    \midrule
    FixMatch & 29.7 & 42.3 \\
    FixMatch+DARP & 30.5 & 42.5\\
    FixMatch+DARP+cRT & 39.7 & 51.0\\
    FixMatch+CReST & 32.5 & 44.7 \\
    FixMatch+CReST+LA & 40.9 & 55.9 \\
    FixMatch+ABC & 46.9 & 56.1 \\
    FixMatch+CoSSL & 43.7 & 53.8 \\
    FixMatch+CDMAD & 48.4 & 59.3 \\
    \rowcolor[rgb]{ .949,  .949,  .949} FixMatch+LCGC & 49.0 & 59.8\\
    \bottomrule
    \end{tabular}
  \caption{Comparison of bACC/GM on Small-ImageNet-127 (size 32 $\times$ 32 and 64 $\times$ 64, $\gamma_u$ is assumed to be known)}
  \label{Tab:small-ImageNet}
\end{table}%

\begin{table*}[!htbp]
  \centering
    \begin{tabular}{cccc}
    \toprule
    \multicolumn{1}{l}{\textbf{Ablation study} ($\gamma_l=100$, $\gamma_u=150$)} & \multicolumn{1}{c}{\textbf{bACC/GM}} & \multicolumn{1}{l}{} & \multicolumn{1}{c}{\textbf{bACC/GM}} \\
    \midrule
    \multicolumn{1}{l}{\textbf{FixMatch+LCGC}} & \multicolumn{1}{c}{\textbf{84.0/83.7}} & \multicolumn{1}{l}{\textbf{ReMixMatch+LCGC}} & \multicolumn{1}{c}{\textbf{83.9/83.6}}\\
    \midrule
    \multicolumn{1}{l}{Without LCGC for refining pseudo-labels} & \multicolumn{1}{c}{81.3/80.4} & \multicolumn{1}{l}{Without LCGC for refining pseudo-labels} & \multicolumn{1}{c}{83.6/83.0}\\
    \multicolumn{1}{l}{Without LCGC for test phase} & \multicolumn{1}{c}{75.9/73.2} & \multicolumn{1}{l}{Without LCGC for test phase} & \multicolumn{1}{c}{78.4/76.5}\\
    \multicolumn{1}{l}{With the use of confidence threshold $\tau = 0.95$} & \multicolumn{1}{c}{83.0/82.6} & \multicolumn{1}{l}{With the use of distribution alignment technique} & \multicolumn{1}{c}{82.1/81.4}\\
    \bottomrule
    \end{tabular}
  \caption{Ablation study for the proposed algorithm on CIFAR-10-LT under $\gamma_l=100$ and $\gamma_u=150$}
  \label{Tab:ablation}
\end{table*}%

Table \ref{Tab:small-ImageNet} summarizes bACC of the baseline algorithms on Small-ImageNet-127. LCGC achieves the best performance for image sizes 32 and 64, respectively. This result highlights the effectiveness of our method in addressing the challenges of CISSL on large-scale and complex datasets.

\subsection{Qualitative Analyses}
\label{ssec:analysis}
In Figure \ref{fig:confusion_matrices} (a) and (b), we observe the class probability distributions under various configurations for FixMatch and ReMixMatch algorithms enhanced with CDMAD and LCGC debiasing techniques. Notably, in Figure \ref{fig:confusion_matrices} (a), LCGC shows significantly higher probabilities for the head classes and lower probabilities for the tail classes than CDMAD, indicating that it is more sensitive to the bias caused by class imbalance. Therefore, in the testing phase, using the logit of the baseline image to refine the over-biased logits can achieve better results than CDMAD.
\begin{figure}[!b]
    \centering
    \includegraphics[width=0.70\columnwidth]{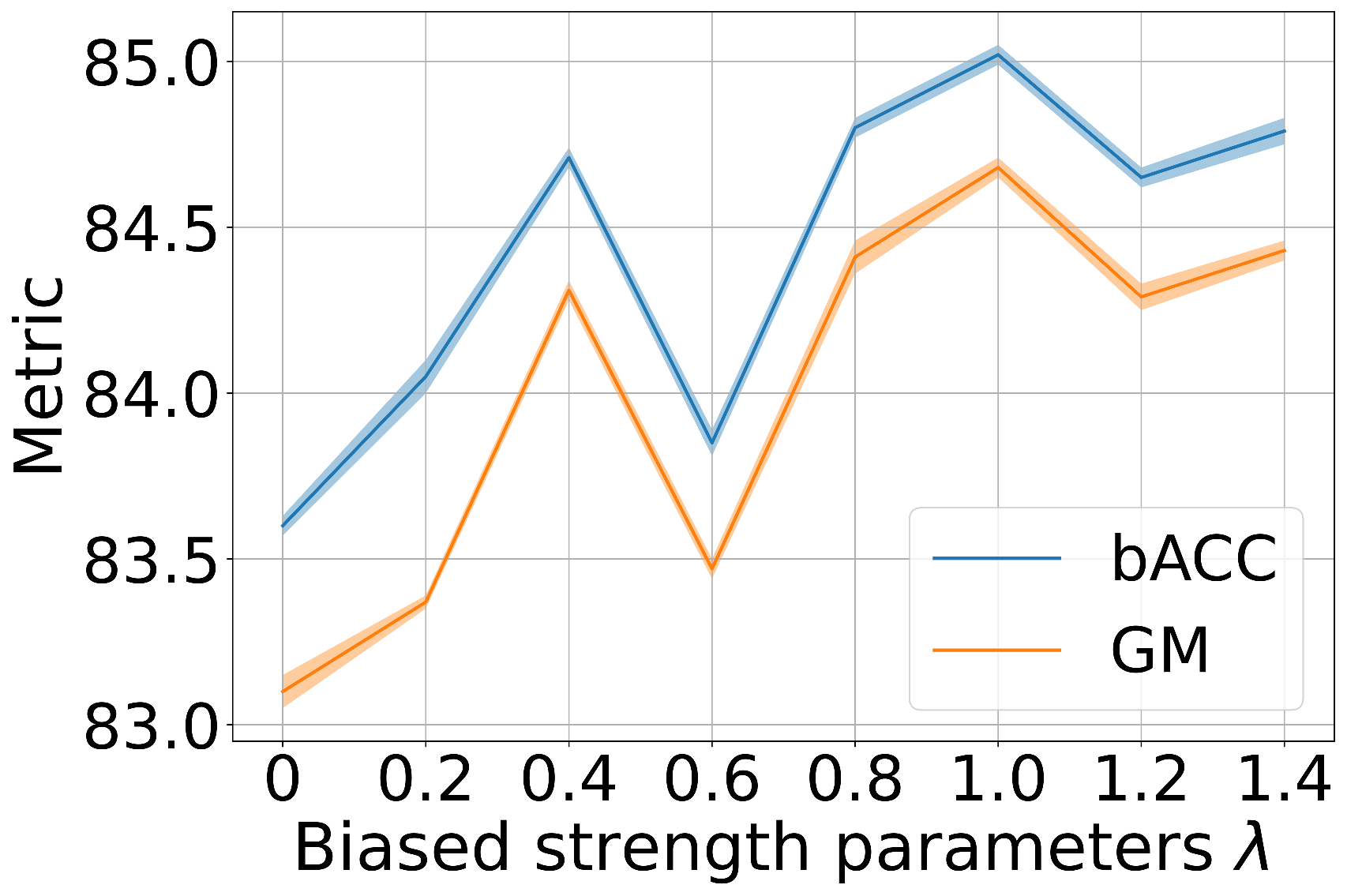}
    \caption{Line chart of validation bACC and GM for the CIFAR-10-LT ($\gamma_l = \gamma_u = 100$) dataset across a range of hyperparameter $\lambda$.}
    \label{fig:parameters}
\end{figure}
Figure \ref{fig:confusion_matrices} (c) and (d) display the confusion matrices of the class predictions on the test set of CIFAR-10 using FixMatch+CDMAD and FixMatch+LCGC trained on CIFAR-10LT under $\gamma_l = 100$ and $\gamma_u = 150$. Notably, the LCGC algorithm demonstrates improved classification accuracies across several classes, particularly in handling minority class samples more effectively.
We conducted an ablation study to examine the impact of each component of LCGC using CIFAR-10-LT with $\gamma_l=100$ and $\gamma_u=150$, as summarized in Table \ref{Tab:ablation}. The results indicate that removing the component for refining pseudo-labels results in a substantial performance drop. Similarly, excluding LCGC during the test phase also negatively impacts performance. Moreover, utilizing a confidence threshold ($\tau = 0.95$) and applying distribution alignment techniques yield slightly lower performance compared to the full LCGC implementation.
\begin{table}[!htbp]
  \centering
    \begin{tabular}{ccc}
    \toprule
    \multicolumn{1}{c}{FixMatch+LCGC} & \multicolumn{2}{c}{CIFAR-10-LT} \\
    \midrule
    Input &   $\gamma_l = \gamma_u = 100$   &  $\gamma_l=100$, $\gamma_u=150$\\
    \midrule
    Red &\multicolumn{1}{c}{83.6/83.2} & \multicolumn{1}{c}{82.1/81.6}\\
    Green &\multicolumn{1}{c}{83.7/83.4} & \multicolumn{1}{c}{82.3/81.9}\\
    Blue &\multicolumn{1}{c}{84.6/84.3} & \multicolumn{1}{c}{83.0/82.6}\\
    Gray &\multicolumn{1}{c}{84.5/84.1} & \multicolumn{1}{c}{83.9/83.4}\\
    White &\multicolumn{1}{c}{84.9/84.6} & \multicolumn{1}{c}{84.0/83.6}\\
    \rowcolor[rgb]{ .949,  .949,  .949} Black &\multicolumn{1}{c}{\textbf{84.9}/\textbf{84.6}} & \multicolumn{1}{c}{\textbf{84.0}/\textbf{83.7}}\\
    \bottomrule
    \end{tabular}
    \caption{Experiments with the replacement of $\mathcal{I}$}
  \label{Tab:replacement}
\end{table}%

\subsection{Sensitivity Analysis}
We hold out the CIFAR-10-LT ($\gamma_l = \gamma_u = 100$) as a validation set, and we report the bACC and GM across a range of parameter values in Figure \ref{fig:parameters}. Note that we choose hyperparameters ranging from 0-1.4 with an interval of 0.2. We select $\lambda = 1.0$ as they correspond to the smallest values.

\subsection{Ablation Study}
\label{sec:ab}
We also conducted experiments to evaluate the effectiveness of replacing the solid color image $\mathcal{I}$ with various color inputs in the LCGC debiasing process using CIFAR-10-LT with $\gamma_l = \gamma_u = 100$ and $\gamma_l=100$, $\gamma_u=150$, as summarized in Tab \ref{Tab:replacement}.  Notably, the use of a black image consistently yielded the highest performance across both $\gamma_l = \gamma_u = 100$ and $\gamma_l=100$, $\gamma_u=150$ settings. The experiments with other color inputs, such as red, green, and blue, showed a slight reduction in performance compared to black. This is due to the Batch Normalization (BN) layer in the classifier backbone: solid color images yield uniform or near-uniform outputs after convolution, which the BN layer standardizes to zero mean and unit variance. As a result, solid black and white images perform similarly (though slight numerical differences exist before rounding), and other solid color images exhibit similar effects. Conversely, noise images produce random pixel distributions that vary across regions, leading to inconsistent outputs after BN standardization and thus poorer performance.  From the analysis, it is evident that the original solid color image, particularly black, plays a critical role in the debiasing process of LCGC.

\section{Conclusion}
\label{sec:con}
We proposed the LCGC algorithm, a novel approach designed to leverage consistency gradient conflicting for improved class representation. We give a theoretical basis for using a baseline image to debias and further integrate gradient guidance information to further improve the debias model. Our comprehensive experiments on benchmark datasets demonstrated that LCGC improves classification accuracy. It is particularly noteworthy that our method is more effective under the class imbalance setting, and the effect becomes better as the imbalance ratio increases.

\section{Acknowledgments}
This work was supported by the Beijing Natural Science Foundation under Grant (No.L231005), and by the National Key Research and Development Program of China under Grant (No.2024YFB3312200). Yue Cheng would like to thank Jiuqian Dai for the useful discussion. 

\bibliography{aaai25}

\begin{thebibliography}{35}
\providecommand{\natexlab}[1]{#1}

\bibitem[{Bengio et~al.(2020)Bengio, Deleu, Rahaman, Ke, Lachapelle, Bilaniuk,
  Goyal, and Pal}]{Bengio2020A}
Bengio, Y.; Deleu, T.; Rahaman, N.; Ke, N.~R.; Lachapelle, S.; Bilaniuk, O.;
  Goyal, A.; and Pal, C. 2020.
\newblock A Meta-Transfer Objective for Learning to Disentangle Causal
  Mechanisms.
\newblock In \emph{International Conference on Learning Representations}.

\bibitem[{Berthelot et~al.(2020)Berthelot, Carlini, Cubuk, Kurakin, Sohn,
  Zhang, and Raffel}]{Berthelot_2020_iclr}
Berthelot, D.; Carlini, N.; Cubuk, E.~D.; Kurakin, A.; Sohn, K.; Zhang, H.; and
  Raffel, C. 2020.
\newblock ReMixMatch: Semi-Supervised Learning with Distribution Matching and
  Augmentation Anchoring.
\newblock In \emph{International Conference on Learning Representations}.

\bibitem[{Cai, Wang, and Hwang(2021)}]{Cai_2021_ICCV}
Cai, J.; Wang, Y.; and Hwang, J.-N. 2021.
\newblock ACE: Ally Complementary Experts for Solving Long-Tailed Recognition
  in One-Shot.
\newblock In \emph{Proceedings of the IEEE/CVF International Conference on
  Computer Vision (ICCV)}, 112--121.

\bibitem[{Cao et~al.(2019)Cao, Wei, Gaidon, Arechiga, and
  Ma}]{NEURIPS2019_621461af}
Cao, K.; Wei, C.; Gaidon, A.; Arechiga, N.; and Ma, T. 2019.
\newblock Learning Imbalanced Datasets with Label-Distribution-Aware Margin
  Loss.
\newblock In Wallach, H.; Larochelle, H.; Beygelzimer, A.; d\textquotesingle
  Alch\'{e}-Buc, F.; Fox, E.; and Garnett, R., eds., \emph{Advances in Neural
  Information Processing Systems}, volume~32. Curran Associates, Inc.

\bibitem[{Chen et~al.(2022)Chen, Jiang, Wang, Wan, Wang, and
  Long}]{NEURIPS2022_d10d6b28}
Chen, B.; Jiang, J.; Wang, X.; Wan, P.; Wang, J.; and Long, M. 2022.
\newblock Debiased Self-Training for Semi-Supervised Learning.
\newblock In \emph{Advances in Neural Information Processing Systems},
  volume~35, 32424--32437.

\bibitem[{Chen et~al.(2023)Chen, Tao, Fan, Wang, Wang, Schiele, Xie, Raj, and
  Savvides}]{chen2023softmatch}
Chen, H.; Tao, R.; Fan, Y.; Wang, Y.; Wang, J.; Schiele, B.; Xie, X.; Raj, B.;
  and Savvides, M. 2023.
\newblock SoftMatch: Addressing the Quantity-Quality Tradeoff in
  Semi-supervised Learning.
\newblock In \emph{The Eleventh International Conference on Learning
  Representations}.

\bibitem[{Cui et~al.(2019)Cui, Jia, Lin, Song, and Belongie}]{Cui_2019_CVPR}
Cui, Y.; Jia, M.; Lin, T.-Y.; Song, Y.; and Belongie, S. 2019.
\newblock Class-balanced loss based on effective number of samples.
\newblock In \emph{Proceedings of the IEEE/CVF Conference on Computer Vision
  and Pattern Recognition (CVPR)}, 9268--9277.

\bibitem[{Du et~al.(2024)Du, Zhang, Zhang, Wu, Wu, and Li}]{DU2024110358}
Du, G.; Zhang, J.; Zhang, N.; Wu, H.; Wu, P.; and Li, S. 2024.
\newblock Semi-supervised imbalanced multi-label classification with label
  propagation.
\newblock \emph{Pattern Recognition}, 150: 110358.

\bibitem[{Fan et~al.(2022)Fan, Dai, Kukleva, and Schiele}]{Fan_2022_CVPR}
Fan, Y.; Dai, D.; Kukleva, A.; and Schiele, B. 2022.
\newblock Cossl: Co-learning of representation and classifier for imbalanced
  semi-supervised learning.
\newblock In \emph{Proceedings of the IEEE/CVF Conference on Computer Vision
  and Pattern Recognition (CVPR)}, 14574--14584.

\bibitem[{Feng et~al.(2024)Feng, Xie, Fang, and Lin}]{Feng_Xie_Fang_Lin_2024}
Feng, Q.; Xie, L.; Fang, S.; and Lin, T. 2024.
\newblock BaCon: Boosting Imbalanced Semi-supervised Learning via Balanced
  Feature-Level Contrastive Learning.
\newblock \emph{Proceedings of the AAAI Conference on Artificial Intelligence},
  38(11): 11970--11978.

\bibitem[{Gidaris, Singh, and Komodakis(2018)}]{gidaris_2018_iclr}
Gidaris, S.; Singh, P.; and Komodakis, N. 2018.
\newblock Unsupervised Representation Learning by Predicting Image Rotations.
\newblock In \emph{International Conference on Learning Representations}.

\bibitem[{Huang et~al.(2016)Huang, Li, Loy, and Tang}]{Huang_2016_CVPR}
Huang, C.; Li, Y.; Loy, C.~C.; and Tang, X. 2016.
\newblock Learning deep representation for imbalanced classification.
\newblock In \emph{Proceedings of the IEEE/CVF Conference on Computer Vision
  and Pattern Recognition (CVPR)}, 5375--5384.

\bibitem[{Huang et~al.(2021)Huang, Xue, Han, Yang, and
  Gong}]{huang2021universal}
Huang, Z.; Xue, C.; Han, B.; Yang, J.; and Gong, C. 2021.
\newblock Universal Semi-Supervised Learning.
\newblock In \emph{Advances in Neural Information Processing Systems}.

\bibitem[{Kim et~al.(2020{\natexlab{a}})Kim, Hur, Park, Yang, Hwang, and
  Shin}]{Kim_2020_nips}
Kim, J.; Hur, Y.; Park, S.; Yang, E.; Hwang, S.~J.; and Shin, J.
  2020{\natexlab{a}}.
\newblock Distribution aligning refinery of pseudo-label for imbalanced
  semi-supervised learning.
\newblock volume~33, 14567--14579.

\bibitem[{Kim et~al.(2020{\natexlab{b}})Kim, Hur, Park, Yang, Hwang, and
  Shin}]{NEURIPS2020_a7968b43}
Kim, J.; Hur, Y.; Park, S.; Yang, E.; Hwang, S.~J.; and Shin, J.
  2020{\natexlab{b}}.
\newblock Distribution Aligning Refinery of Pseudo-label for Imbalanced
  Semi-supervised Learning.
\newblock In Larochelle, H.; Ranzato, M.; Hadsell, R.; Balcan, M.; and Lin, H.,
  eds., \emph{Advances in Neural Information Processing Systems}, volume~33,
  14567--14579. Curran Associates, Inc.

\bibitem[{Kubat, Matwin et~al.(1997)}]{Kubat_1997_Icml}
Kubat, M.; Matwin, S.; et~al. 1997.
\newblock Addressing the curse of imbalanced training sets: one-sided
  selection.
\newblock In \emph{Icml}, volume~97, 179. Citeseer.

\bibitem[{Lai et~al.(2022)Lai, Wang, Gunawan, Cheung, and
  Chuah}]{pmlr-v162-lai22b}
Lai, Z.; Wang, C.; Gunawan, H.; Cheung, S.-C.~S.; and Chuah, C.-N. 2022.
\newblock Smoothed Adaptive Weighting for Imbalanced Semi-Supervised Learning:
  Improve Reliability Against Unknown Distribution Data.
\newblock In \emph{Proceedings of the 39th International Conference on Machine
  Learning}, volume 162, 11828--11843.

\bibitem[{Lee and Kim(2024)}]{Lee_2024_CVPR}
Lee, H.; and Kim, H. 2024.
\newblock CDMAD: Class-Distribution-Mismatch-Aware Debiasing for
  Class-Imbalanced Semi-Supervised Learning.
\newblock In \emph{Proceedings of the IEEE/CVF Conference on Computer Vision
  and Pattern Recognition (CVPR)}, 23891--23900.

\bibitem[{Lee, Shin, and Kim(2021)}]{NEURIPS2021_3953630d}
Lee, H.; Shin, S.; and Kim, H. 2021.
\newblock ABC: Auxiliary Balanced Classifier for Class-imbalanced
  Semi-supervised Learning.
\newblock In \emph{Advances in Neural Information Processing Systems},
  volume~34, 7082--7094.

\bibitem[{Li et~al.(2024)Li, Tao, Han, Zhan, and Ye}]{Li_Tao_Han_Zhan_Ye_2024}
Li, L.; Tao, B.; Han, L.; Zhan, D.-c.; and Ye, H.-j. 2024.
\newblock Twice Class Bias Correction for Imbalanced Semi-supervised Learning.
\newblock \emph{Proceedings of the AAAI Conference on Artificial Intelligence},
  38(12): 13563--13571.

\bibitem[{liang et~al.(2021)liang, Wu, Li, Wang, Meng, Qin, Chen, Zhang, and
  Liu}]{NEURIPS2021_5a66b920}
liang, x.; Wu, L.; Li, J.; Wang, Y.; Meng, Q.; Qin, T.; Chen, W.; Zhang, M.;
  and Liu, T.-Y. 2021.
\newblock R-Drop: Regularized Dropout for Neural Networks.
\newblock In \emph{Advances in Neural Information Processing Systems},
  volume~34, 10890--10905.

\bibitem[{Miyato et~al.(2018)Miyato, Maeda, Koyama, and
  Ishii}]{miyato2018virtual}
Miyato, T.; Maeda, S.-i.; Koyama, M.; and Ishii, S. 2018.
\newblock Virtual adversarial training: a regularization method for supervised
  and semi-supervised learning.
\newblock \emph{IEEE transactions on pattern analysis and machine
  intelligence}, 41(8): 1979--1993.

\bibitem[{Nam et~al.(2020)Nam, Cha, Ahn, Lee, and Shin}]{NEURIPS2020_eddc3427}
Nam, J.; Cha, H.; Ahn, S.; Lee, J.; and Shin, J. 2020.
\newblock Learning from Failure: De-biasing Classifier from Biased Classifier.
\newblock In \emph{Advances in Neural Information Processing Systems},
  volume~33, 20673--20684.

\bibitem[{Oh, Kim, and Kweon(2022)}]{Oh_2022_CVPR}
Oh, Y.; Kim, D.-J.; and Kweon, I.~S. 2022.
\newblock DASO: Distribution-Aware Semantics-Oriented Pseudo-Label for
  Imbalanced Semi-Supervised Learning.
\newblock In \emph{Proceedings of the IEEE/CVF Conference on Computer Vision
  and Pattern Recognition (CVPR)}, 9786--9796.

\bibitem[{Schmutz, HUMBERT, and Mattei(2023)}]{schmutz2023dont}
Schmutz, H.; HUMBERT, O.; and Mattei, P.-A. 2023.
\newblock Don{\textquoteright}t fear the unlabelled: safe semi-supervised
  learning via debiasing.
\newblock In \emph{The Eleventh International Conference on Learning
  Representations}.

\bibitem[{Sohn et~al.(2020)Sohn, Berthelot, Carlini, Zhang, Zhang, Raffel,
  Cubuk, Kurakin, and Li}]{Sohn_2020_nips}
Sohn, K.; Berthelot, D.; Carlini, N.; Zhang, Z.; Zhang, H.; Raffel, C.~A.;
  Cubuk, E.~D.; Kurakin, A.; and Li, C.-L. 2020.
\newblock FixMatch: Simplifying Semi-Supervised Learning with Consistency and
  Confidence.
\newblock In \emph{Advances in Neural Information Processing Systems},
  596--608.

\bibitem[{Sundararajan, Taly, and Yan(2017)}]{pmlr_2017_icml}
Sundararajan, M.; Taly, A.; and Yan, Q. 2017.
\newblock Axiomatic Attribution for Deep Networks.
\newblock In \emph{Proceedings of the 34th International Conference on Machine
  Learning}, 3319--3328.

\bibitem[{Tao et~al.(2024)Tao, Li, Li, and Zhan}]{10446881}
Tao, B.; Li, L.; Li, X.-C.; and Zhan, D.-C. 2024.
\newblock CLAF: Contrastive Learning with Augmented Features for Imbalanced
  Semi-Supervised Learning.
\newblock In \emph{ICASSP 2024 - 2024 IEEE International Conference on
  Acoustics, Speech and Signal Processing (ICASSP)}, 7185--7189.

\bibitem[{Wang et~al.(2023{\natexlab{a}})Wang, Jia, Wang, Wu, and
  Meng}]{wang2023imbalanced}
Wang, R.; Jia, X.; Wang, Q.; Wu, Y.; and Meng, D. 2023{\natexlab{a}}.
\newblock Imbalanced Semi-supervised Learning with Bias Adaptive Classifier.
\newblock In \emph{The Eleventh International Conference on Learning
  Representations}.

\bibitem[{Wang et~al.(2022)Wang, Wu, Lian, and Yu}]{Wang_2022_CVPR}
Wang, X.; Wu, Z.; Lian, L.; and Yu, S.~X. 2022.
\newblock Debiased Learning From Naturally Imbalanced Pseudo-Labels.
\newblock In \emph{Proceedings of the IEEE/CVF Conference on Computer Vision
  and Pattern Recognition (CVPR)}, 14647--14657.

\bibitem[{Wang et~al.(2023{\natexlab{b}})Wang, Xu, Yang, He, Cao, and
  Huang}]{NEURIPS2023_973a0f50}
Wang, Z.; Xu, Q.; Yang, Z.; He, Y.; Cao, X.; and Huang, Q. 2023{\natexlab{b}}.
\newblock A Unified Generalization Analysis of Re-Weighting and
  Logit-Adjustment for Imbalanced Learning.
\newblock In Oh, A.; Naumann, T.; Globerson, A.; Saenko, K.; Hardt, M.; and
  Levine, S., eds., \emph{Advances in Neural Information Processing Systems},
  volume~36, 48417--48430. Curran Associates, Inc.

\bibitem[{Wei et~al.(2021)Wei, Sohn, Mellina, Yuille, and Yang}]{wei2021crest}
Wei, C.; Sohn, K.; Mellina, C.; Yuille, A.; and Yang, F. 2021.
\newblock Crest: A class-rebalancing self-training framework for imbalanced
  semi-supervised learning.
\newblock In \emph{Proceedings of the IEEE/CVF conference on computer vision
  and pattern recognition}, 10857--10866.

\bibitem[{Wei and Gan(2023)}]{Wei_2023_CVPR}
Wei, T.; and Gan, K. 2023.
\newblock Towards Realistic Long-Tailed Semi-Supervised Learning: Consistency
  Is All You Need.
\newblock In \emph{Proceedings of the IEEE/CVF Conference on Computer Vision
  and Pattern Recognition (CVPR)}, 3469--3478.

\bibitem[{Zhang et~al.(2021)Zhang, Cui, Xu, Zhou, He, and
  Shen}]{Zhang_2021_CVPR}
Zhang, X.; Cui, P.; Xu, R.; Zhou, L.; He, Y.; and Shen, Z. 2021.
\newblock Deep Stable Learning for Out-of-Distribution Generalization.
\newblock In \emph{Proceedings of the IEEE/CVF Conference on Computer Vision
  and Pattern Recognition (CVPR)}, 5372--5382.

\bibitem[{Zhu et~al.(2024)Zhu, Liu, Wu, Tang, Tang, Niu, and
  Su}]{zhu2024estimating}
Zhu, G.; Liu, X.; Wu, X.; Tang, S.; Tang, C.; Niu, J.; and Su, H. 2024.
\newblock Estimating before Debiasing: A Bayesian Approach to Detaching Prior
  Bias in Federated Semi-Supervised Learning.
\newblock \emph{arXiv preprint arXiv:2405.19789}.

\end{thebibliography}

\end{document}